\documentclass[]{article}
\usepackage{enumerate}
\usepackage{listings}
\usepackage{latexsym,bm,amsmath,amssymb}
\usepackage{amsthm}
\usepackage{graphicx}
\usepackage{mathrsfs}
\usepackage[]{hyperref}
\usepackage{booktabs}
\usepackage{xcolor}
\usepackage{bm}
\usepackage{framed}
\usepackage{extarrows}
\usepackage{multirow}
\usepackage{authblk}

\usepackage[verbose=true,letterpaper]{geometry}
\AtBeginDocument{
  \newgeometry{
    textheight=9in,
    textwidth=6.5in,
    top=1in,
    headheight=14pt,
    headsep=25pt,
    footskip=30pt
  }
}

\begin{document}
\title{Next Waves in Veridical Network Embedding}

\author[1,**]{Owen G.\ Ward}
\author[1,**]{Zhen Huang}
\author{Andrew Davison}
\author[1,2]{Tian Zheng\thanks{Correspondence should be sent to tian.zheng@columbia.edu.}}
\affil[1]{Department of Statistics, Columbia University, New York, NY.}
\affil[2]{Data Science Institute, Columbia University, New York, NY.}
\affil[**]{These authors contributed equally.}
\date{July 2020}








\maketitle


\abstract{Embedding nodes of a large network into a metric (e.g., Euclidean) space has become an area of active research in statistical machine learning, which has found applications in natural and social sciences. Generally, a representation of a network object is learned in a Euclidean geometry and is then used for subsequent tasks regarding the nodes and/or edges of the network, such as community detection, node classification and link prediction. Network embedding algorithms have been proposed in multiple disciplines, often with domain-specific notations and details. In addition, different measures and tools have been adopted to evaluate and compare the methods proposed under different settings, often dependent of the downstream tasks. As a result, it is challenging to study these algorithms in the literature systematically. Motivated by the recently proposed 
PCS framework for
Veridical Data Science, we propose a framework for network embedding algorithms and discuss how the principles of \emph{predictability}, \emph{computability} and \emph{stability} (PCS)
apply in this context. The utilization of this framework in network embedding holds the potential to motivate and point to new directions for future research.}


\section{Introduction}
Complex relationships amongst a collection of objects can be encoded in 
network data. A major challenge with data of this form is that 
it is difficult to directly use this data structure for certain analysis 
that provides insight into the properties of the network, such as for 
clustering nodes, community detection, and predicting future links between
nodes.
For example, in social networks,
there may be interest in
identifying whether two users are friends 
\cite{backstrom2011supervised,grover2016node2vec}, or which product 
a user would be interested in \cite{perozzi2014deepwalk}. In linguistic 
networks, semantic and syntactic relationships between different words 
can be of interest \cite{mikolov2013linguistic}. 
In biological networks,
we may wish
to identify relationships between a specific gene and certain
symptoms or diseases or
the existence of interaction between proteins 
\cite{vazquez2003global,grover2016node2vec}, while in 
chemical networks we may wish to predict certain 
properties of a molecule \cite{gilmer2017neural}.
Most methods which aim to address these tasks, including most commonly 
used machine learning methods, generally require data consisting of 
features (i.e., vectors) in Euclidean space. This allows the many 
well developed procedures for statistical learning in Euclidean geometry, such as gradient based optimization, to be applied. Thus, methods which can fully represent this network structure in a well understood geometry 
are crucial to providing further analysis of network data, and it is an active area of research in both the statistics and machine learning 
communities.

Many promising methods have been proposed so far, some of which
take ideas from language models.
For example, Deepwalk  \cite{perozzi2014deepwalk} and node2vec \cite{grover2016node2vec} regard nodes as \emph{words} and random walks as \emph{sentences} to learn the low dimensional representations by SkipGram, a Natural Language Processing (NLP) model.
Such approaches generate interesting results but may be
worrisome in that the way words compose together to make 
a sentence is not necessarily the same as
how nodes connect in a graph.
The latter may contain richer graph structural information.

A defining feature of all network embedding
methods is the subsequent analysis which
will be performed using the learned representation.
This is crucial in analysing the performance
of a method. As such, one needs to identify
the features of the network which will be
required for this analysis, and which should
be retained by the embedding process.

In this paper, we summarise the
problem of network embeddings for
a statistical audience, detailing the
key features of any network embedding
method. We illustrate and discuss some of the
most commonly used algorithms.
We then discuss network embedding in the
context of the \emph{Veridical
Data Science} principles 
recently proposed in  \cite{yu2020veridical}.
Finally, we consider some of the open
problems in network embedding
for the statistical community,
including the use of more general
distributions of network features,
the opportunities offered by embeddings
in non-Euclidean spaces, and the
need for more rigorous evaluation
and comparison of the embedding
methods proposed.


\section{Problem Setup}
A network is usually represented by $G=(V,E)$,
where $V$ is a set of $n$ vertices or nodes, and $E$ is the set of edges
between these vertices, 
together with an adjacency matrix $A$, with $A_{ij}$ indicating the 
presence or absence of an edge between nodes $i$ and $j$.
The edges between nodes can be weighted or unweighted,
where, if there is a weight associated with an edge between nodes
$i$ and $j$, it is denoted as $W_{ij}$, an element of a weight 
matrix $W$.
It is also possible for these edges to be either
directed or undirected. In general, it is assumed that
there are no self edges.
Similarly, it is possible that there is also some other $m$
covariates associated with each node, which can
be expressed as a vector $x_i \in \mathcal{X} \subseteq \mathbb{R}^m$. Edge features are also possible and can be similarly defined.

The goal of network embedding is to map the information present
in the network to some representation space,
$\mathcal{Z}$, which is assumed to be a metric space
with associated metric $\rho$. That is, we wish
to construct a mapping 
\begin{equation}
    \Phi: G\times \mathcal{X} \to (\mathcal{Z},\rho).
\end{equation}

When constructing a network embedding, many aspects of the problem 
need to be considered. 
\begin{itemize}
    \item Which space do we wish to represent the network
in?
    \item  What features of the network do we require this embedding to 
preserve and how do we evaluate whether these features are preserved? 
    \item If we had nodal covariates present in the network, how do we best combine them with
the learned representation? 
\item Given that we have constructed an embedding of our network, what task
will this embedding then be used for?
\end{itemize}
We discuss each of these concerns below.


\subsection*{Representation space}
When performing a network embedding, the first question
which must be addressed is the space $\mathcal{Z}$ which the network
will be embedded in.
The most popular and practical choice is possibly $\mathcal{Z}=\mathbb{R}^d$.
It is generally assumed that $d \ll n$, in which case this
will also lead to a lower dimensional representation of the network.
This dimension reduction may lead
to improved computational
performance if this representation
is then used in further tasks.

While machine learning methods have exclusively used Euclidean
geometry for network embedding, in the context of statistical
latent space models, other geometries 
have been considered.
Other representation spaces such as the $d$-dimensional unit sphere $S^d$  \cite{mccormick2015latent}, and the hyperbolic space $\mathbb{H}^d$ are also possible. The hyperbolic space, for example, may capture some tree-like structure of certain networks in real life.
See Smith et al \cite{smith2019geometry} for a review of the different latent space geometries used in practice.

\subsection*{Features of network to be preserved}
Given that we have chosen a space in which we wish to embed our
network, we must then decide what features present in the network
we want this embedding to preserve, and how we evaluate the
preservation of these features in our embedding space. 
The structure of the network we wish the embedding to reflect
can also be crucial
for later tasks using the learned representation.
One such example is community detection. An embedding which
will be used for community detection should require that
if there is a \emph{community structure}, based on a quantifiable definition, present in the network,
the resulting embeddings of nodes
should reflect this. 
This can be challenging when there are multiple community structures present, such
as different subsets of nodes displaying distinct community structure.
Nodes in the same community 
should be embedded closer to each other than nodes in different 
communities, leading to an effective clustering solution.

Many embeddings seek to preserve a form of ``proximity" between
nodes to capture similarity between these nodes. Proximity can be captured
by both the presence of edges between nodes and also by the weights
of these edges (if applicable). The \emph{first-order proximity}, for example, is commonly used to capture direct similarity between nodes that share an edge.
If there is an edge from node $i$ to node $j$ with a large weight $W_{ij}$, then node $i$ and node $j$ are regarded to have high first-order proximity.
If the task is clustering, and it is believed that people who communicate frequently tend to belong to the same group, then the \emph{first-order proximity} may be a good metric to capture this  \cite{belkin2003laplacian,ahmed2013distributed,tang2015line,wang2016structural}.

Some other common features in the literature are \emph{second-order 
proximity} \cite{tang2015line,wang2016structural}, \emph{$k$-step transition 
probability} \cite{cao2015grarep}, \emph{homophily} \cite{grover2016node2vec}, structural 
equivalence \cite{grover2016node2vec}, and features which capture other additional
information associated with nodes such as attributes of nodes \cite{hamilton2017inductive,scarselli2008graph}, 
text \cite{yang2015network}, or labels \cite{scarselli2008graph}. 
For example, the \emph{second-order proximity} measures neighborhood 
differences. In a social network, it explores the idea that if two people have many friends 
in common, then they may be friends as well, even without the presence 
of a direct edge between them. 
The \emph{$k$-step transition 
probability} 
regards the 
weight on edge $W_{ij}$ as an unnormalized transition probability
in the context of a random walk on nodes of the network. If 
the current state of a random walk is at node $i$,
then the random walk will
jump to node $j$
with probability $\frac{W_{ij}}{\sum_k W_{ik}}$ at the next step.
A high transition probability from node 
$i$ to $j$ can be an indication that they are ``close" in some
sense, which then indicates that they should also be similar
in the embedding space.
There is also \emph{homophily},
which requires that people from the same community
should be 
embedded closely to each other, along
with structural equivalence, requiring that nodes with alike structural 
roles
(e.g. the center of a community or a hub node) should be embedded 
similarly \cite{grover2016node2vec}.
The local neighborhood structure is also explored by some models in a 
variety of ways. For example, a neural network model can be set up to 
aggregate information from neighboring nodes at each 
layer \cite{gilmer2017neural,hamilton2017inductive,li2015gated}.

Another important question is how best to incorporate nodal 
covariates
into either learning the representation or being used with the
representation for further tasks. For example, covariates 
could
be used directly to learn the representation 
 \cite{kipf2016semi},
as is commonly done in neural network embedding models
 \cite{hamilton2017inductive,li2015gated,velivckovic2018deep}.
It seems natural for 
embedding methods to also preserve measures
of proximity defined on these nodal covariates, of which a detailed
review is given in  \cite{cui2018survey}.


\subsection*{Similarity in the representation space}
Having chosen an embedding for the network in the representation
space we can then consider what similarity is present in this 
space.
Most embeddings are constructed based on the idea that ``similarity" in 
the network should be represented by a ``similarity" in the 
representation space.
While there are many choices of network similarity as discussed above, 
similarity in the representation space
has been less explored. In general, where the embedding space
is a metric space, then the metric distance between the embedding of 
nodes provides an automatic similarity. 
In Euclidean geometry, the related (but not equivalent)
dot product is also commonly used, due to computational convenience.
Alternative similarity measures include \emph{cosine} similarity and other variants based on angular separation \cite{hoff2002latent}.
These similarities are often used on the unit sphere, where the
cosine similarity $s(z_i,z_j)=\frac{z_i^T z_j}{\|z_i\|\cdot \|z_j\|}$ coincides with the dot product up to rescaling, and the angular similarity $s(z_i,z_j)=\arccos\left({\frac{z_i^T z_j}{\|z_i\|\cdot \|z_j\|}}\right)$ coincides with the spherical distance.

\section{Representational learning of networks}

\paragraph{Learning the representation}
Given a choice of embedding space, the network features we 
wish to preserve, and the concept of similarity in the embedding space,
we must then consider the procedure for 
learning this embedding.
Several procedures have been 
considered in learning the representation of a network. We first
describe some of the most commonly used, before discussing the evaluation
of these learning methods.

For unsupervised representation learning, most methods are
developed based on the idea of ``matching the 
similarities" of the network and the space. That is, nodes
that are ``close" in the original network should also be 
``close" in the representation space.
One approach is to assume that there is some generating process
from the representations to the features of the network or vice
versa. The variables to be optimized can be any parameters in the
generating process including the representation itself.
The generating process can be deterministic, such as using a 
neural
network \cite{wang2016structural,hamilton2017inductive}. The learning can 
then be carried out by 
minimizing the construction error or matching the similarities. 
For example, SDNE \cite{wang2016structural} uses a deep auto-encoder with 
the rows of the graph adjacency matrix $A$ (or the weight matrix for weighted graphs) as inputs. The loss function is
$$\begin{aligned}
\mathcal{L}&=\mathcal{L}_2+\alpha\mathcal{L}_1+\nu \mathcal{L}_{reg}\\
&=\|(\hat{A}-A)\odot B\|^2_F + \alpha\sum_{i,j=1}^n A_{i,j}\|z_i-z_j\|_2^2 + \frac{\nu}{2}\sum_{k=1}^K(\|W^{(k)}\|^2_F+\|\hat{W}^{(k)}\|_F^2),
\end{aligned}$$
where $z$ is the representation, and $\hat{A}$ is the reconstruction of input $A$ based on $z$. $\|\cdot\|_F$ is the Frobenius norm of matrix. $W^{(k)}$ and $\hat{W}^{(k)}$ are the weights of the encoding and decoding neural networks.
$\mathcal{L}_1$ matches the \emph{first-order proximity} of the two spaces.
$\mathcal{L}_{reg}$ is for regularization, and
$\mathcal{L}_2$ measures the reconstruction error weighted by $B$, with $B_{ij}=1$ if $A_{ij}=0$ and $B_{ij}=\beta>1$ if $A_{ij}>0$.
$\mathcal{L}_2$ seeks to preserve the second-order proximity, i.e. the neighborhood similarity, since the row of the adjacency matrix can be viewed as a characterization of a vertex's neighbors.

An explicit generating process is not always required.
One alternative is to directly consider two similarity 
measures, $S(v_i,v_j)$ between nodes in the network, and
$S'(z_i,z_j)$ between their representations in the
representation space. We can then define a loss function
based on these similarities, with one natural choice
being
$$L(S,S')=\sum_{i,j}S(v_i,v_j)S'(z_i,z_j)+R(z),$$
for some constraint or 
regularization term on the embedding, $R(z)$. For example,
in the Laplacian eigenmap method \cite{belkin2003laplacian}, 
$S(v_i,v_j)=W_{ij}$ is 
the \emph{first-order proximity}, $S'(z_i,z_j)=\|z_i-z_j\|_2^2$, 
$R(z)$ is some normalizing constraints on $z$, and $L$ is to be 
minimized. Similarly, in GraRep \cite{cao2015grarep}, $S(v_i,v_j)=p_k(v_j|v_i)$ is taken as the
$k$-step transition probability and 
$S'(\Phi(v_i),\Phi(v_j))=\log\sigma(z_j'^Tz_i)$, where 
$\sigma(x)=\frac{1}{1+\exp(-x)}$, and $\Phi(v_i)=(z_i,z_i')$ 
which is associated with two vectors.
$z_i'$ is the representation when $v_i$ is viewed as ``context'' (more details are given
in Section \ref{Deepwalk}).
In this case $L$ is then maximized. 
Most unsupervised methods are of this form, minimizing some reconstruction
loss between the network and the representation.
We summarize some unsupervised methods, along with their key
properties, in Table~\ref{tab:methods_similarities}.

\begin{table}[ht]
 \caption{Some unsupervised methods, their similarity measures
    and the corresponding loss function used to learn the representation. 
    Here
    $S(v_i,v_j)$ is the similarity on the network, $S'(z_i,z_j)$ is the 
    similarity in the representation space, $z_i$ is the representation of 
    $v_i$, and $W$ is the weight matrix.}
  \label{tab:methods_similarities}
  \begin{center}
   \renewcommand{\arraystretch}{1.3}
    \begin{tabular}{llll}
    \hline
      {\bf Methods} & $S(v_i,v_j)$ & $S'(z_i,z_j)$ & {\bf Loss Function}\\
      \hline
      Laplacian eigenmap \cite{belkin2003laplacian}&$W_{ij}$&$\|z_i-z_j\|^2$&$\sum_{i,j}S(v_i,v_j)S'(z_i,z_j), \quad {\rm s.t.}\ \mathbf{z}^TD\mathbf{z}=I.$ \\
      \hline
      Graph factorization \cite{ahmed2013distributed}& $W_{ij}$&$z_i^Tz_j$ &$\frac{1}{2}\sum_{(i,j)\in E}(S(v_i,v_j)-S'(z_i,z_j))^2 +\frac{\lambda}{2}\sum_i \|z_i\|^2$ \\
      \hline
      LINE(1st-Order) \cite{tang2015line}&$W_{ij}$&$\log\sigma(z_i^Tz_j)$&$-\sum_{i,j} S(v_i,v_j)S'(z_i,z_j)$\\
      \hline
      LINE(2nd-Order) \cite{tang2015line}&$W_{ij}$&
      $\frac{\exp(z_j'^Tz_i)}{\sum_k\exp(z_k'^Tz_i)}$&$-\sum_{i,j}S(v_i,v_j)S'(z_i,z_j)$\\
      \hline
      GraRep \cite{cao2015grarep}&$p_k(v_j|v_i)$&$\log\sigma(z_j'^Tz_i)$&
      $\sum_{i,j}^{|V|}\big[ S(v_i,v_j)S'(z_i,z_j) +\frac{\lambda}{|V|}\mathbb{E}_{v_k\sim p_k}\log[1-e^{S'(z_i,z_k)}]\big]$\\ 
      \hline
      Deepwalk \cite{perozzi2014deepwalk} & co-occurrence & $\log(\sigma(z_{v_j}'^T z_{v_i}))$ & $\log(\sigma(z_{v_j}'^T z_{v_i})) +\sum_{l=1}^k\mathbb{E}_{v_l\sim
    P_n}\log(1-\sigma(z_{v_l}'^Tz_{v_i}))$ \\
      Node2vec \cite{grover2016node2vec} & in RWs & & \\
      \hline
    \end{tabular}
  \end{center}
\end{table}

If the generating process is probabilistic, parameterized by the 
representations of the nodes, then the representations can be 
learnt by maximizing the likelihood function. 
Deepwalk \cite{perozzi2014deepwalk} and Node2vec \cite{grover2016node2vec}, for example,
maximize the co-occurrence probability of nodes appearing in a 
random walk across the network. A fully generative probabilistic model is also possible. Smith et
al.  \cite{smith2019geometry} reviewed a number of parametric models of the 
form:
$$\begin{aligned}
    P({\rm edge\ between}\ (v_i,v_j))&\overset{\rm 
    indep}{\sim}{\rm Bernoulli}(p_{ij}),\quad i\neq j,\\
    {\rm logit}(p_{ij}&)=\alpha + s(z_i,z_j),\\
    z_i &\overset{i.i.d.}{\sim} \mathbb{P}(f|\psi),
\end{aligned}$$
where $z_i=\Phi(v_i)$ is the representation of node $v_i$, and 
$s(\cdot,\cdot)$ is some similarity measure such as distance or 
dot product. We discuss latent space models of this form
in more detail below.

For supervised and semi-supervised learning, the learnt representations can be
directed towards specific tasks, and are usually able to achieve better 
performances on prediction than the unsupervised methods.
In such cases, a downstream process can immediately take the representations 
as input to perform certain task of interest, along with a task-specific loss 
to be optimized. For example, Tu et al. \cite{tu2016max} applied support 
vector machines (SVM) to the representations learnt by Deepwalk for 
multi-label classification. The version of Deepwalk based on matrix 
factorization was used. Apart from the matrix factorization loss, the SVM 
classification loss on those labeled nodes was also added, which results in a 
mix of supervised and unsupervised loss:
$$\begin{aligned}
\min_{Z,Y,W,\xi}\mathcal{L}&=\min_{Z,Y,W,\xi} \mathcal{L}_{DW} + \frac{1}{2}\|W\|_2^2+C\sum_{i=1}^T\xi_i,\\
{\rm s.t.}\ w_{l_i}^Tz_i - w_j^Tz_i&\geq 1_{l_i\neq j} - \xi_i,\ \forall i,j,\\
{\rm where}\ \mathcal{L}_{DW}&=\|M-Z^TY\|^2_2 +\frac{\lambda}{2}(\|Z\|_2^2+\|Y\|_2^2).
\end{aligned}$$

A purely supervised loss is also possible.
There is a rich literature on deep learning on networks (see  \cite{zhang2018deep} for a detailed review). Despite the many different architectures that are adopted, many of the neural network models are equipped with a supervised loss.
The network takes the graph and the associated node/edge features as input, and produces certain output for the task of interest. The error between the output and the true response is then optimized to learn the network \cite{scarselli2008graph,li2015gated,gilmer2017neural,kipf2016semi}. In the \emph{Message Passing Neural Networks} \cite{gilmer2017neural} framework which unifies a number of earlier methods, the last hidden states (layer) before output can be taken as the representations of the nodes.

The optimization of the loss function, after possible
approximations and relaxations, is often carried out by (stochastic) gradient descent \cite{bottou1991stochastic}, and sometimes can be run in parallel if the update is sparse \cite{recht2011hogwild,perozzi2014deepwalk,grover2016node2vec}.
If the final optimization problem is convex, then techniques in convex optimization can be applied \cite{tu2016max}.
In some cases, matrix factorization can also be used for optimizing the loss function \cite{cao2015grarep},
and recent work has attempted to rephrase common random walk
methods under
the framework of matrix factorization  \cite{qiu2018network},
which could lead to better interpretation of these
methods. However,
this matrix factorization approach cannot
readily
scale to large networks,
and as such is not of current 
computational benefit.


\paragraph{Assessing the representation}
Having constructed our representation we can then assess its performance.
Assessment of the representation can be in terms of reconstruction of the 
original network and also the subsequent inference formed based on the
embedding  \cite{cui2018survey}.
For downstream inference using the network representation, common
tasks include node classification, link prediction and node 
clustering.
These tasks often present further decisions before the final
analysis is complete. For example, to perform node clustering 
using
a learned representation, the clustering method to be used and 
the
number of clusters to be considered must also be specified. 
Similarly,
if the goal is link prediction then a sample of the observed 
links are 
treated as the test set and removed, with the embedding fit on 
the
remaining edges  \cite{lu2011link}. 
As such, when evaluating the overall method on these downstream 
tasks alone,
it is difficult to precisely quantify the effect of the specific 
embedding chosen. It is thus important to specify which goal is
prioritized as part of the network embedding when determining the
method
to be used. The PCS framework provides a natural
workflow to address these concerns.

\section{Review of representative methods}

\subsection*{Methods from statistics}
\paragraph{Spectral Clustering}
One of the most well studied models for 
the clustering on nodes in a network,
in the
statistical literature, is spectral clustering.
This implicitly constructs an embedding of the network using 
a collection of eigenvectors from the spectral
decomposition of the graph laplacian. The eigenvectors of the graph laplacian
corresponding to the $k$ smallest eigenvalues are then used to construct
the matrix $U\in \mathbb{R}^{n\times k}$, where the $i$th row can be thought of
as a $k$ dimensional representation of the $i$th node.
A distance-based clustering
method, such
as $k$-means clustering,
is then applied to the representation of these
nodes in $\mathbb{R}^k$.

Spectral clustering has been widely studied in the 
statistical literature and as such is one of the few network
embedding models for which theoretical properties are 
known, including consistent recovery of communities in networks 
 \cite{lei2015consistency},
and of an underlying latent space  \cite{rubin2017statistical},
along with more general consistency results
 \cite{von2008consistency}. However, it has also been shown
to not be robust in the presence of outliers 
 \cite{cai2015robust}. Moreover, spectral clustering is not
always computationally feasible for large networks, requiring an expensive
singular value decomposition, though it can be accelerated using \emph{locally optimal block preconditioned conjugate gradient} \cite{knyazev23toward}, since only the eigenvectors corresponding to the smallest eigenvalues are needed. 
We summarize this and other methods we will
discuss further
using ideas from model cards \cite{mitchell2019model} in Appendix \ref{model_cards}.

\paragraph{Latent Space Models}
Another model which is commonly considered to 
represent networks in the statistical community are
latent space models for social networks \cite{hoff2002latent}, of which 
the stochastic 
block model \cite{nowicki2001estimation} is a popular special
case. Latent space models of this form posit that the true
network given by an adjacency matrix
can be represented in some lower dimensional latent space
and then propose estimation procedures to learn the positions
of the nodes in this latent space.
These differ somewhat from other network embeddings, which
just aim to represent the network in some latent space. This model and extensions, which include
clustering in the 
latent space,
learning both the latent space and clusters in that 
space \cite{handcock2007model},
have been fit using both MCMC and variational 
inference methods 
 \cite{hoff2002latent,
salter-townshend_variational_2013}.
These inference procedures also inhibit scaling this
method to larger networks, due to the expensive likelihood
computation required for parametric models of this form,
although approximations have been 
developed  \cite{raftery2012fast}.

\subsection*{Methods from machine learning}

\paragraph{Random-walk based network embedding}
\label{Deepwalk}
Motivated by the advance of natural language processing (NLP), a number of network embedding algorithms have been developed in recent years.
Popular models include DeepWalk \cite{perozzi2014deepwalk} and Node2vec \cite{grover2016node2vec}.
The main idea is to view each node in the graph as a word in the corpus.
Sentences in NLP correspond to random walks on 
a graph which are generated according to certain rules.
The word embedding methods in NLP can then 
be applied to learn the network embedding.

Different methods have been used to generate random
walks exploring different features of a network.
DeepWalk \cite{perozzi2014deepwalk} and GraRep \cite{cao2015grarep} do this by directly using the network edge weight.
The weights on edge $W_{ij}$ are regarded as unnormalized transition probabilities. If the current state is at node $i$, then the random walk goes to node $j$ with probability $\frac{W_{ij}}{\sum_k W_{ik}}$ at the next step,
which is basically a finite-state Markov model.
Node2vec \cite{grover2016node2vec} modified the above transition 
probability by introducing two additional hyperparameters, which can 
adjust the random walk between Breadth-first Sampling and Depth-first 
Sampling,
and thus become a second-order Markov chain. 
After obtaining the random walks, certain objective function is optimized.
For example, DeepWalk applies SkipGram \cite{mikolov2013distributed} with 
gradient descent steps to minimize $-\log{\rm 
Pr}\left(\{v_{i-w},\cdots,v_{i-1},v_{i+1},\cdots,v_{i+w}\}|\Phi(v_i) 
\right)$, where $\Phi(v_i)$ is the representation of node $v_i$, for each 
vertex $v_i$ in the random walk $(v_1,\cdots,v_L)$ and a pre-specified 
window size $w$.
A natural and common choice for ${\rm Pr}\left(\cdot|\Phi(v_i) \right)$ is
to use the softmax function, which assigns independently, for each $v_j$,
$$
    {\rm Pr}(v_j|\Phi(v_i))=\frac{e^{z_{v_j}'^Tz_{v_i}}}{\sum_{v_k\in 
    V}e^{z_{v_k}'^Tz_{v_i}}},
$$
where $\Phi(v_i)=(z_{v_i},z'_{v_i})$. Each vertex has two roles, either 
being the ``centered node" or the ``context node" within the local window 
of other ``centered node". So two different vectors $z_i$, $z_i'$ are 
associated respectively.
The ``centered node" representation $z_{v_i}$ is usually taken as the 
final representation for output.

However, objectives involving softmax function are not computationally 
feasible for large networks, since calculating the denominator takes 
$O(|V|)$ time.
One approximation is to use
the \emph{hierarchical softmax} approximation
\cite{morin2005hierarchical,perozzi2014deepwalk}.
It builds a binary tree with all 
the nodes as its leaves.
With an input $z_{v_i}$ at the root, the probability ${\rm 
Pr}(v_j|\Phi(v_i))$ is calculated along the path from the root to 
the leaf of $v_j$, reducing the computational complexity to 
$O(\log|V|)$.

Another popular approximation is
\emph{negative sampling} \cite{mikolov2013efficient,grover2016node2vec,cao2015grarep}.
It approximates $\log{\rm Pr}(v_j|\Phi(v_i))$ by
\begin{equation}
    \log(\sigma(z_{v_j}'^T z_{v_i})) +\sum_{l=1}^k\mathbb{E}_{v_l\sim 
    P_n}\log\sigma(-z_{v_l}'^Tz_{v_i}),
    \label{simplified_loss}
\end{equation}
where $\sigma(x)=\frac{1}{1+\exp(-x)}$ and $P_n$ is some noise 
distribution, positing a
good model should be able to 
distinguish data from noise.
In practice, $P_n$ is usually taken as the unigram distribution raised to 3/4 power:
$P_n(v_i)\propto [{\rm frequency\ of\ }v_i]^{3/4}$.
Though theoretical guarantees do 
not exist, it has been found that this choice outperforms several other candidates significantly, such as the unigram and the uniform distributions \cite{mikolov2013distributed}.
Corresponding experiments for network embeddings are still absent according to our knowledge. The objective \eqref{simplified_loss} is then maximized by stochastic 
gradient descent \cite{bottou1991stochastic}.
At each iteration, $v_l,(l=1,\cdots,k)$ are sampled from $P_n$, and then a 
usual gradient ascent step is performed.
When the window size is small,
each gradient step only updates a small portion of the 
embedding vectors. Parallel optimization can also be made possible and has
been used in practice  
\cite{recht2011hogwild,perozzi2014deepwalk,grover2016node2vec}.

\paragraph{Neural Network Models}
There is a rich literature on neural network methods 
on graphs (see  \cite{zhang2018deep} for a detailed 
review).
The \emph{graph neural network} 
(GNN) \cite{scarselli2008graph} was first proposed in a
recursive way:
$$\begin{aligned}
z_n=f_w(l_n,l_{{\rm co}[n]},z_{{\rm ne}[n]},l_{{\rm 
ne}[n]}),\quad 
o_n=g_w(z_n,l_n),
\end{aligned}$$
where $z_n$, $l_n$, $l_{{\rm co}[n]}$, $z_{{\rm 
ne}[n]}$,
$l_{{\rm ne}[n]}$, $o_n$ are the representation of 
node $n$, the label of $n$, the labels of edges with 
$n$ being one vertex, the representations of nodes in 
the neighborhood of $n$, the labels of nodes in the 
neighborhood of $n$, and 
the output, respectively. $f_w$ and
$g_w$ are some 
parametric functions. Each step of the 
semi-supervised learning process is carried out by 
first iterating the equation to the unique fixed 
point, and then performing a gradient descent to 
minimize the error between the output and the true 
response of the supervised nodes. One limitation of 
this model is that in order for the iteration scheme
to give a (unique) fixed point, $f_w$ is essentially required
to be a contraction mapping with respect to a complete metric (from which the contraction mapping theorem guarantees the existence and uniqueness of a fixed point). 
Also, iterating the equation to 
convergence can be computationally expensive.
Many neural network models have been
proposed without 
this recursive definition, but with different 
convolutional network structures.
Some spectral methods use the eigenvectors of the 
graph Laplacian to construct convolution 
operators \cite{bruna2013spectral}.
Others may utilize the graph structure to aggregate 
information from neighbors at each layer to update the
representations
 \cite{hamilton2017inductive,li2015gated,gilmer2017neural}.
\emph{Message Passing Neural Networks} 
(MPNN) \cite{gilmer2017neural} unifies
a number of 
earlier proposed convolutional neural network models. Its update is of
the following form at the $(t+1)^{\rm th}$ 
layer:
$$\begin{aligned}
m_v^{t+1}&=\sum_{w\in N(v)}M_t(z_v^t,z_w^t,e_{vw}),\\
z_v^{t+1}&=U_t(z_v^t,m_v^{t+1}),\\
\hat{y}&=R(\{z_v^T|v\in G\}).
\end{aligned}$$
where $M_t$, $N(v)$, $z_v^t$, $e_{vw}$, $U_t$, 
$\hat{y}$, $R$ are learnable message functions, 
neighbors of node $v$, the representation of $v$ at 
time $t$, edge feature, learnable vertex update 
functions, the output, a learnable readout function, 
respectively.
Some neural network embedding methods, by design, can 
be generalized to unseen nodes, which is called \emph{inductive} learning \cite{hamilton2017inductive}.
In that case, $N(v)$ can be taken as a fixed size of neighbors, uniformly sampled from all neighbors of $v$.
The counterpart is \emph{transductive} 
learning \cite{hamilton2017inductive}, which requires 
all nodes to be present during model fitting.

Most of the above mentioned neural network models are 
supervised or semi-supervised, with a task-specific 
loss function minimizing the error between the outputs
and the targets. There are also unsupervised deep 
learning methods such as deep autoencoder, 
GraphSAGE \cite{hamilton2017inductive}, and Deep Graph 
Infomax \cite{velivckovic2018deep}.
Deep autoencoder methods, e.g. 
SDNE \cite{wang2016structural}, try to minimize the 
reconstruction error. 
GraphSAGE \cite{hamilton2017inductive} uses the 
following loss function on the final representations 
$z_v,v\in V$:
$$J_G(z_v)=-\log(\sigma(z_v^Tz_u))-Q\mathbb{E}_{v_n
\sim P_n}\log(\sigma(-z_v^Tz_{v_n})),$$
where $u$ is a node that co-occurs near $v$ on 
fixed-length random walk, $\sigma$ is the sigmoid 
function, $P_n$ is a negative sampling distribution, 
and $Q$ is the number of negative samples.
Deep Graph Infomax \cite{velivckovic2018deep} is a 
recently proposed unsupervised method which 
outperforms previous unsupervised methods, along
with some 
supervised models,
on a number of benchmark datasets. It
obtains the representations by a function 
$\mathcal{E}$, which is usually a learnable neural 
network, acting on the features $\mathbf{X}$ and the 
graph adjacency matrix $\mathbf{A}$, 
with representation 
$\mathbf{Z}=\mathcal{E}(\mathbf{X},\mathbf 
A)=(z_1,\cdots,z_N)$.
The objective function aims to maximize the mutual 
information of $z_i$ and $s=\mathcal{R}(\mathbf{Z})$, 
a graph-level summary statistic, which can be taken as
$\mathcal{R}(\mathbf{Z})=\sigma\left(\frac{1}{N}\sum_{i=1}^N z_i\right)$ in practice. The objective 
function is:
$$\mathcal{L}=\frac{1}{N+M}\left(\sum_{i=1}^N\mathbb{E}_{(\mathbf{X},\mathbf{A})}\left[\log\mathcal{D}\left(z_i,s\right)\right]+\sum_{j=1}^M\mathbb{E}_{(\tilde{\mathbf{X}},\tilde{\mathbf{A}})}\left[\log\left(1-\mathcal{D}\left(\tilde{z}_j,s\right)\right)\right] \right),$$
where $(\tilde{\mathbf{X}},\tilde{\mathbf{A}})=C(X,A)$
are the negative samples obtained by the (stochastic) 
corruption function $C$, and $\tilde{\mathbf{Z}}=(\tilde{z}_1,\cdots,\tilde{z}_M)=\mathcal{E}(\tilde{\mathbf{X}},\tilde{\mathbf{A}})$ are
the representations of negative samples.
$\mathcal{D}$ is a discriminator which models the probability of the pairs $(z_i,s)$, and can be taken as $\mathcal{D}(z_i,s)=\sigma\left(z_i^T W s\right)$ in practice, with $W$ a learnable matrix. 

Having considered this procedure, and
the procedures previously discussed, some
further questions naturally arise. In particular, for this method,
is this algorithm stable under 
different but reasonable
choices of the corruption function $C$ or the encoder 
$\mathcal{E}(\mathbf{X},\mathbf{A})$?
More generally,
will the above methods work well on other networks of 
different sizes or from different domains? Is the 
computation required for each of these methods
still feasible as the size of the network increases?
It is desirable to address questions of this form
in a systematic manner. This is
a natural motivation to consider a framework for 
doing this. We will attempt to address these questions
using the PCS framework for veridical
data science\cite{yu2020veridical}.

\begin{figure}[ht]
    \centering
    \includegraphics[width = 1\textwidth]{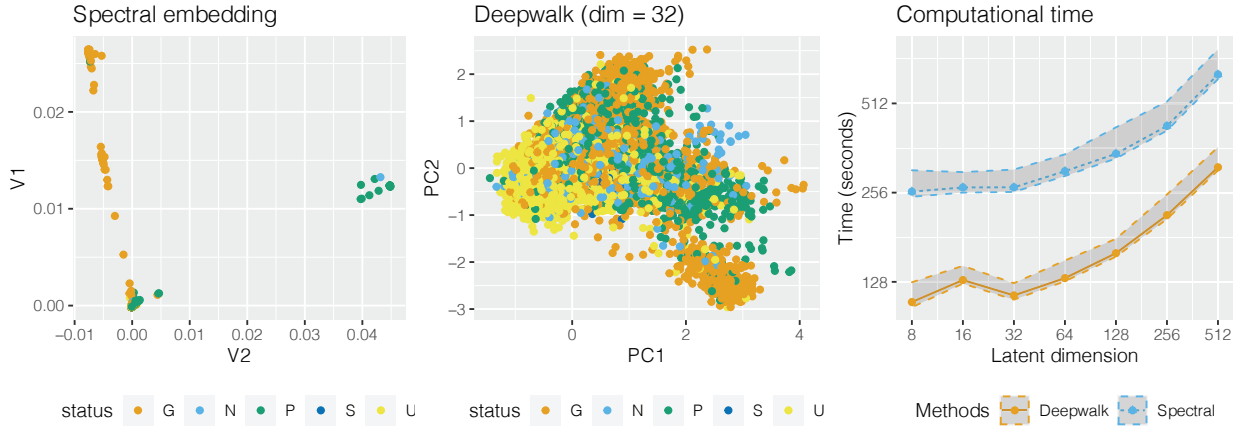}
    \caption{ Left and middle: the 2-dimensional projection of the embeddings of student email network produced by spectral embedding and Deepwalk (with latent dimension $d=32$). The spectral embedding is plotted using the first two eigenvectors, while the Deepwalk
    embedding
    is plotted using the first two principal components. Student's status include U (undergrad), G (graduate), P (professional), N (non-degree), and S (statement of attendance). Right: the mean computation time of the embeddings, averaged over 100 replications. The confidence bands show the 2.5\% and 97.5\% sample quantiles of the computational times.
}\label{embeds2D}
\end{figure}

\begin{figure}[ht]
    \centering
    \includegraphics[width = \textwidth]{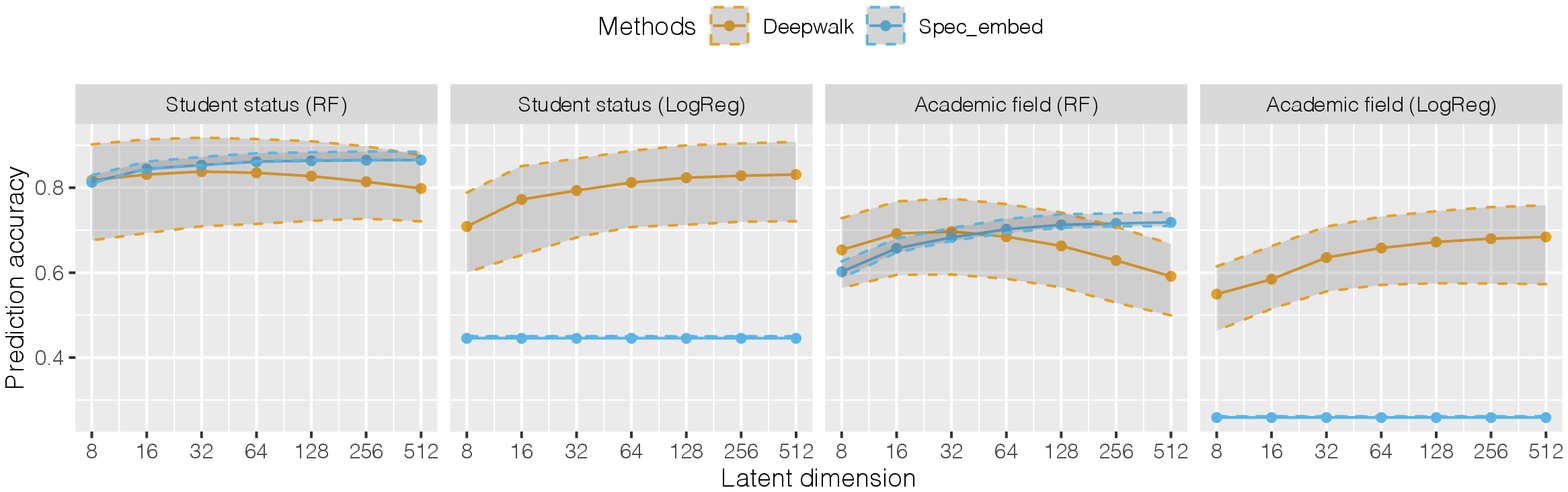}
    \caption{ The mean accuracy of prediction (averaged over 100 replications, each with 5 scores from stratified cross-validation) for one of the 12 academic fields and one of the 5 student status with different machine learning models (random forest and logistic regression). The confidence bands are plotted using the 2.5\% and 97.5\% sample quantiles.}\label{two_tasks}
\end{figure}

\paragraph{An illustrative example}
To further illustrate the properties of some of these popular 
methods, we apply them to a large real network dataset.
Here we consider an email network 
of all students and staff
at a US university during one semester 
in the academic year 2003-2004. This data was first analysed by
Kossinets and Watts
\cite{kossinets2006empirical}. We consider here the network
of all students at the university, with available nodal information
available.
We treat this network as 
unweighted and undirected, with an edge existing between two students if 
they communicate via email during the semester. By taking the largest 
connected component, we obtain a graph with 18492 nodes and 260048 edges.
We apply spectral embedding (implemented in scikit-learn 
\cite{pedregosa2011scikit}) and Deepwalk \cite{perozzi2014deepwalk} to 
learn the latent representations, both with default settings.
Figure~\ref{embeds2D} shows the 
embeddings projected to a plane. The spectral 
embedding is visualized using the first two eigenvectors,
while the Deepwalk 
is visualized
using the first two principal components from a 32-dimensional 
latent representation.
The computational costs (in seconds, run on the
same machine with 4 cores in
parallel) are given in the right-hand side of Figure~\ref{embeds2D}.
The embeddings are then used to perform prediction on the academic fields (12 categories) and the student status (5 categories) for each student. Stratified 5-fold cross-validated accuracies are then computed for random forest classifier and logistic regression. Both classifiers were implemented in 
scikit-learn in default settings. A larger maximum iteration number is set for logistic regression to ensure convergence.
Figure~\ref{two_tasks} shows the accuracy of prediction for the two 
methods on the test set.
Since there is randomness in the random-walk as well as the training/test 
sets splitting, we repeat the whole procedure 100 times, with confidence 
bands showing the 2.5\% and 97.5\% sample quantiles.

From this systematic assessment, we can see that: (a) for latent dimension $d = 16, 
32$, spectral embedding is better at predicting student status, but worse at 
predicting academic field compared with Deepwalk; (b) Deepwalk is computationally 
faster, but spectral embedding is more stable; (c) even for the same predicting task 
(predicting student status), different machine learning models (random forest and 
logistic regression) may provide different conclusions about the two methods. When 
using a random forest classifier, spectral embeddings achieve higher accuracy, while 
Deepwalk performs better when a logistic regression is used. 
Even this relatively simple network embedding task illustrates some of the components
involved in constructing an embedding of a network and using this representation
for further tasks. In the following section, we will discuss these
components in the context of \cite{yu2020veridical}.

\section{A Veridical Network Embedding?}
The principles of predictability, computability and stability,
as introduced in  \cite{yu2020veridical}, can be used to construct
a framework for veridical data science (VDS), and this
framework is of clear benefit in the context of 
network embedding. We will discuss each of these principles in
turn and comment on their role in the context of
evaluating network embedding models.

\emph{Predictability}, as defined in the PCS framework,
seeks to construct a
simple metric to evaluate how well a model represents relationships
in the original data. This is in terms of some prediction target
from the data, which may be observed or extracted (supervised
or unsupervised). Many papers describing network embedding models
consider metrics of this form. For example, link prediction on
held out edges \cite{grover2016node2vec} or classification of nodes
with known labels \cite{perozzi2014deepwalk}. The held out test set
used for this evaluation should not differ from the original data.

We illustrate such a node classification task using the previously
described embeddings shown in Figure~\ref{embeds2D}.
In Figure~\ref{two_tasks}
we see that the performance of these embeddings varies for 
the two nodal classifications considered, and that this
performance also changes with the downstream classification
tool used on these learned embeddings, highlighting
the multiple aspects of the embedding process that need to
be considering for an eventual prediction task.

One important concern when link prediction is used is the difficulty 
in 
sampling
edges of a network to remove, and the theoretical properties of 
network
sampling are still somewhat lacking. 
Recall that in a traditional classification setting where data are
assumed to be drawn independently and identically from some distribution, 
we form 
training and test splits in order to estimate the generalization error of our 
classifier
(or use multiple  training test splits 
and cross-validation in order to estimate the expected test error).
These splits are created via sub-sampling the original data
in a way which `respects' the way the 
observed data are sampled from a hypothetical population distribution.
In the case of 
networks, various candidates for population models of graphs (including  
\cite{lovasz_large_2012, veitch_class_2015, veitch_sampling_2016, 
borgs_sampling_2017, borgs_sparse_2018,caron_sparse_2017, Crane2018})
have been proposed,
each with their own shortcomings. It is therefore 
unclear if in practice one should accept a single routine way in which training 
splits should be chosen, although some work
\cite{orbanz2017subsampling} has explored the 
implications of when
a subsampling procedure defined on finite graphs leads to a suitable population 
limit.
These concerns are important when considering tasks that involve sampling a 
network.

This is closely related to the implementation of validation methods, such
as cross validation for network models.
While some such methods have recently been proposed \cite{li2016network}, they
have not
yet been applied to evaluate network embedding models, and further
work is needed to better understand these procedures on networks.

\emph{Computability}, which in the PCS framework 
is described as an issue of algorithm
efficiency and scalability, is of clear importance in network 
embedding.
Many network embedding procedures seek to embed large networks 
containing hundreds of thousands of nodes, 
and the use of simpler low
dimensional representations naturally aids computation.
However, there is of course a tradeoff between computation and 
the representation power of the learned embedding; larger embedding 
dimensions allow for more flexible representations but come at the 
cost at more expensive computation.
Approximations are commonly used to perform this embedding, such as 
the
widely used negative sampling 
\cite{perozzi2014deepwalk,grover2016node2vec} and the hierarchical 
softmax appromimation  \cite{perozzi2014deepwalk}.
These methods induce optimization objectives which are amenable to
stochastic gradient descent methods, and allow for the
use of minibatches and the possibility of distributed updates, 
although there are the previously discussed
concerns with subsampling network data.
As these optimization procedures are non-convex, there are also 
questions as to 
whether the learned embeddings are
global minima of the objectives from which they are learned. 
Scalability also hinders (more severely) other methods such as 
spectral clustering 
and MCMC based methods such as latent space models, limiting the use 
of these models to networks with at most tens of thousands of 
vertices. 
We see in Figure~\ref{embeds2D} that, for our illustrative 
email
network, spectral clustering is more computationally expensive
that the Deepwalk embedding method for all choices of latent 
dimension,
with both methods becoming more
expensive as the latent dimension increases.

\emph{Stability} seeks to identify how the result of an analysis changes when
the data and/or the model are perturbed. As described in  \cite{yu2020veridical},
this is required across the network embedding procedure, and should
take account of all components of the network embedding problem: 
\begin{itemize}
    \item The choice of representation space; for instance, do we use
    a Euclidean or hyperbolic geometry for the representation space?
    What embedding dimension do we use?
    \item The features of the network to be preserved; do we preserve
    only first order connectivity information, or higher orders? 
    \item Similarity in the representation space; this can be both 
    from the choice of loss function used when training, and also any
    prescribed similarity measures we use for analysis after 
    training.
    \item The procedure for learning the representation space; this 
    is on the level of whether we use e.g spectral clustering or 
    Deepwalk, and then also for a particular algorithm the choice of 
    hyperparameters (e.g for Deepwalk, this includes the sampling 
    scheme and any other hyperparameters such as lengths of random 
    walks, window sizes, and the step size/initialization chosen for the gradient 
    descent procedure).
    \item The subsequent analysis for which the learned embedded is 
    used; for instance, are we interested in link prediction tasks or
    community detection?
    \item Changes to the data; what happens if a small number of 
    edges are removed from the network? Does the result of our 
    analysis significantly change if some noise is added to the 
    network (by e.g flipping the labels of a small number of edges?)
\end{itemize}

We see some aspects of this stability
component in Figure~\ref{two_tasks}. For example,
as we increase the latent dimension of our embedding space
we see that predicted classifications from a spectral embedding
are stable, with limited improvement from increased dimension.
The embedding learned by Deepwalk, however, appears to
vary, sometimes leading to improved classification accuracy
and sometimes to worse accuracy as the dimension increases.
Similarly, we see that for predicting student status,
using the Deepwalk embedding is stable under different classification methods used on the embedding,
unlike the spectral embedding.
While perturbations of any one component of this procedure
can be evaluated, a metric is needed to fully assess
the stability of models of this form, which shall be able to capture 
perturbations in the data along with both stages of modeling present 
in network embedding (learning the representation and the subsequent 
analysis using this representation)

\section{Discussion}
Network embedding is an exciting area of active research and new approaches
are regularly proposed in the literature. In this paper we have given
a brief review of the problem of network embedding, discussed the
key components of the task, and reviewed in detail some
common procedures from both the statistical and machine
learning fields. 

We have discussed a framework for the evaluation of
network embedding methods, using the principles proposed in
 \cite{yu2020veridical}. 
The goal of such a framework is to provide
``responsible, reliable, reproducible and transparent
results across the data science life cycle"  \cite{yu2020veridical},
something we believe is crucial,
particularly in the context of network 
embedding.
We consider each of the principles 
of predictability, computability, and stability
in this context. We feel that the
utilization of a framework such as this can help to direct
future work on network embedding and help to
better understand the broad existing literature.

There are still many unresolved questions
in this literature which provide further 
opportunities. Embedding richer
network features, which may be needed to address 
domain specific questions  \cite{backstrom2006group},
is still an open problem. Although applying NLP methods directly
to networks has yielded empirical success,
a deeper understanding of these models requires
further work on the interaction
between sampling procedures on networks,
and the impact this sampling has on 
the learning of the embedding. Moreover, there is a question
of interpreting which latent structures of the network
are being recovered by such schemes, and how they may differ from those produced e.g by spectral clustering. Given the interaction between sampling and censoring, this should also lead to insights for link prediction tasks (which we can think of as a missing data problem).
It is hoped that 
utilization of this framework could help 
identify empirical evidence 
to help address these challenges. 
Another question which has been considered 
is the geometry of the
embedding space used and the result structure this can
capture \cite{smith2019geometry}. 
Further work could explore the relationship between embedding
geometries and the PCS framework.

While most existing procedures employ a similar
problem setup, many different outcomes
are considered in the literature, leading to a wide
array of evaluation metrics. As such, a systematic
evaluation of competing methods is required to
better understand the state of the art
and direct further research. The agreement 
on benchmark datasets and metrics, extending beyond link prediction,
would also be required. Extensive simulation studies,
based on the principles proposed here for network embedding,
would help to identify these benchmarks
and metrics, and would undoubtedly
identify further directions of
interest in the study of network representation.
Similarly, while we have exclusively considered the problem of static networks here, there is
also much recent research into dynamic networks which can evolve over time \cite{du2018dynamic},
\cite{nguyen2018continuous},\cite{zhou2018dynamic}. 
There are many differences when considering how to represent a network with a temporal
component, and there are also a wide range of possible evaluation metrics which can
be considered for networks of this form. 
Addressing the concepts of PCS in the development
of embedding methods for these networks is an area
with many future challenges.

\newpage
\appendix
\section{Model cards}\label{model_cards}
\begin{framed}
    \begin{center}
        \textbf{Spectral Clustering of the Graph Laplacian}
    \end{center}
    \begin{itemize}
        \item \textbf{Model Description.} 
        \item \textbf{Inputs.} Adjacency matrix $A$ of a 
        graph $G(V, E)$
        \item \textbf{Output.} Cluster assignments of the nodes in
        the network.
        \item \textbf{Procedure.}
        Given a Laplacian matrix $L$ of $A$, construct  the
        bottom $k$ eigenvectors of $L$, $u_1,\ldots,u_k$ which
        give the matrix $U=(u_1,\ldots,u_k)\in\mathbb{R}^{n\times k}$. 
        Perform a distance clustering using the rows of $U$,
        to cluster each node.
        \item \textbf{Task.} Clustering of the nodes.
        \item \textbf{Source Code.} Implemented in most software packages.
    \end{itemize}
\end{framed}

\begin{framed}
    \begin{center}
        \textbf{Latent Space Models}
    \end{center}
    \begin{itemize}
        \item \textbf{Model Description.} 
        \item \textbf{Inputs.} Adjacency matrix $A$ of a 
        graph $G(V, E)$
        \item \textbf{Output.} Latent position of the nodes in some
        lower dimensional space.
        \item \textbf{Loss Function.}
        Posit a latent space. Log likelihood,
        conditional on that latent space, of the form
        $$
        \sum_{i\neq j} \eta_{i,j}A_{i,j} - 
        \log\left(1 + e^{\eta_{i,j}}  \right),
        $$
        where $\eta_{i,j}$ is the log odds of the probability 
        of an edge between nodes $i$ and $j$, with
        $\eta_{i,j}= \alpha + d_{i,j}$, with $d_{i,j}$ some metric
        distance in the latent space.
        \item \textbf{Task.} Represent nodes in low dimensional
        Euclidean space.
        \item \textbf{Source Code.} Implemented, for example,
        in \texttt{latentnet}
        package in \texttt{R}.
    \end{itemize}
\end{framed}

\begin{framed}
    \begin{center}
        \textbf{DeepWalk \cite{perozzi2014deepwalk}}
    \end{center}
    \begin{itemize}
        \item \textbf{Model Description.} Apply SkipGram to learn network embedding with random walks.
        \item \textbf{Inputs.} Graph $G(V, E)$, window
        size $w$, embedding dimension $d$, number of 
        walks per vertex $\gamma$, walk length $t$.
        \item \textbf{Output.} The vertex 
        representations $\Phi(V)\in 
        \mathbb{R}^{|V|\times d}$.
        \item \textbf{Loss Function.}
        $$\sum_{\mathcal{W}_k}\sum_{v_i\in\mathcal{W}_
        k}-\log{\rm Pr}\left(\{v_{i-w},\cdots,v_{i-1},v_{i+1},
        \cdots,v_{i+w}\}|\Phi(v_i) \right).$$
        \item \textbf{Task.} Multi-Label 
        Classification.
        \begin{itemize}
            \item \textbf{Evaluation Data.} 
            BlogCatalog, Flickr, YouTube.
            \item \textbf{Classification Model.} 
            One-vs-rest logistic regression.
            \item \textbf{Metrics.} Macro-$F_1$ and 
            Micro-$F_1$ scores.
        \end{itemize}
        \item \textbf{Source Code.} 
        \url{https://github.com/phanein/deepwalk}
    \end{itemize}
\end{framed}

\begin{framed}
    \begin{center}
        \textbf{Deep Graph Infomax \cite{velivckovic2018deep}}
    \end{center}
    \begin{itemize}
        \item \textbf{Model Description.} An unsupervised embedding method maximizing mutual infomation.
        \item \textbf{Inputs.} Graph (with features) $\mathbf{A}$, $\mathbf{X}$, corruption function $C$, encoder $\mathcal{E}(\mathbf{X}, \mathbf{A})$, readout function $\mathcal{R}(\mathbf{Z})$, discriminator $\mathcal{D}$.
        \item \textbf{Output.} The vertex representations $\mathbf{Z}=(z_1,\cdots,z_N)$.
        \item \textbf{Loss Function.}
        $$\mathcal{L}=\frac{1}{N+M}\left(\sum_{i=1}^N\mathbb{E}_{(\mathbf{X},\mathbf{A})}\left[\log\mathcal{D}\left(z_i,s\right)\right]+\sum_{j=1}^M\mathbb{E}_{(\tilde{\mathbf{X}},\tilde{\mathbf{A}})}\left[\log\left(1-\mathcal{D}\left(\tilde{z}_j,s\right)\right)\right] \right),$$
        \item \textbf{Tasks.} Multi-Label Classification, visualization of clustering.
        \begin{itemize}
            \item \textbf{Evaluation Data.} \\
            Transductive: Cora, Citeseer, Pubmed;\\
            Inductive: Reddit, PPI
            \item \textbf{Classification Model.} Logistic regression.
            \item \textbf{Metrics.} Micro-$F_1$ scores on test nodes (unseen nodes for inductive tasks). 
        \end{itemize}
        \item \textbf{Source Code.} \url{https://github.com/PetarV-/DGI}
    \end{itemize}
\end{framed}

\newpage
\bibliographystyle{plain}
\bibliography{bibliography}
\end{document}